\documentclass[runningheads]{llncs}
\usepackage[T1]{fontenc}
\usepackage{graphicx}
\usepackage[colorlinks,
            linkcolor=black,
            anchorcolor=green,
            citecolor=blue
            ]{hyperref}
\usepackage{amsfonts}
\usepackage{bm}
\usepackage{amsmath}
\usepackage{autobreak}
\usepackage{multirow}
\usepackage{marvosym}
\usepackage{verbatim}
\usepackage{amssymb}
\usepackage{booktabs}
\usepackage{siunitx}   
\usepackage[table]{xcolor}
\usepackage{array}
\usepackage[acronym]{glossaries}
\usepackage{float}
\usepackage{overpic}
\usepackage{colortbl,xcolor}

\newacronym{ssl}{SSL}{self-supervised learning}
\newacronym{dp}{DP}{digital pathology}
\newacronym{fm}{FM}{foundation model}
\newacronym[plural=WSIs,firstplural=whole-slide images (WSIs)]{wsi}{WSI}{whole-slide image}
\newacronym[plural=RoIs,firstplural=regions of interest (RoIs)]{roi}{RoI}{region of interest}

\newcommand{\std}[1]{{\tiny\ensuremath{\pm #1}}}

\begin{document}
\title{Revisiting Automatic Data Curation for Vision Foundation Models in Digital Pathology}
\titlerunning{Automatic Data Curation in Digital Pathology}
%
\makeatletter
\renewcommand{\@fnsymbol}[1]{%
  \ifcase#1 \or *\or \dag\or \ddag\or \S\or \P\or \|\or **\or \dag\dag\or \ddag\ddag \else \@ctrerr\fi}
\makeatother

\author{Boqi Chen\thanks{Equal contribution}\inst{1}\inst{2}\and
Cédric Vincent-Cuaz\inst{*}\inst{3}\and
Lydia A. Schoenpflug\inst{*}\inst{4} \and
Manuel Madeira\inst{*}\inst{3}\and
Lisa Fournier\inst{6} \and
Vaishnavi Subramanian\inst{3} \and
Sonali Andani\inst{1}\inst{4}\inst{5}\and
Samuel Ruiperez-Campillo\inst{1}\inst{2} \and
Julia E. Vogt\inst{1}\inst{2}\and
Raphaëlle Luisier\inst{6}\and
Dorina Thanou\inst{3}\and
Viktor H. Koelzer\thanks{Co-corresponding authors: \email{viktor.koelzer@usb.ch; pascal.frossard@epfl.ch; gabriele.campanella@mssm.edu; raetsch@ethz.ch}}\inst{4}\inst{5}\and
Pascal Frossard\inst{\dag}\inst{3}\and
Gabriele Campanella\inst{\dag}\inst{7}\and
Gunnar Rätsch\inst{\dag}\inst{1}\inst{2}}
\authorrunning{B. Chen et al.}
\institute{Dept. of Computer Science, ETH Zurich, Zurich, Switzerland\and AI Center, ETH Zurich, Zurich, Switzerland\and
Signal Processing Laboratory (LTS4), EPFL, Lausanne, Switzerland\and
University of Basel, Basel, Switzerland\and
University Hospital of Basel, Basel, Switzerland\and
Idiap Research Institute, Martigny, Switzerland\and
Icahn School of Medicine at Mount Sinai, New York, United States\\
}
\maketitle              
%
%
\begin{abstract}
Vision foundation models (FMs) are accelerating the development of digital pathology algorithms and transforming biomedical research.
These models learn, in a self-supervised manner, to represent histological features in highly heterogeneous tiles extracted from whole-slide images (WSIs) of real-world patient samples. The performance of these FMs is significantly influenced by the size, diversity, and balance of the pre-training data. Yet, data selection has been primarily guided by expert knowledge at the WSI level, focusing on factors such as disease classification and tissue types, while largely overlooking the granular details available at the tile level.
In this paper, we investigate the potential of unsupervised automatic data curation at the tile-level, taking into account 350 million tiles. Specifically, we apply hierarchical clustering trees to pre-extracted tile embeddings, allowing us to sample balanced datasets uniformly across the embedding space of the pretrained FM. We further show that these datasets are subject to a trade-off between size and balance, potentially compromising the quality of representations learned by FMs. We propose tailored batch sampling strategies to mitigate this effect. We demonstrate the effectiveness of our method through improved performance on a diverse range of clinically relevant downstream tasks.

\end{abstract}
%
%
\section{Introduction}
Large-scale pre-trained \gls{ssl} models, or \glspl{fm}, have demonstrated a remarkable ability to learn task-agnostic representations from unlabeled image data, capturing rich domain-specific knowledge~\cite{caron2021emerging,he2022masked}. Their exposure to vast and diverse data sources enables them to learn highly transferable representations for various downstream tasks~\cite{vo2024automatic}. This property is particularly attractive for \gls{dp}, which involves analyzing high-resolution \glspl{wsi} with significant heterogeneity across different biological scales to assess clinically relevant tasks, such as cancer typing and grading, survival prediction and treatment response assessment.

Data curation plays a crucial role in training \glspl{fm}, leading to improved performance~\cite{oquab2023dinov2}. Strategies such as data pruning~\cite{toneva2018empirical,sorscher2022beyond}, active learning~\cite{geifman2017deep,vo2022active}, and nearest neighbor search~\cite{oquab2023dinov2} have proven effective. Recently, clustering-based methods have emerged as competitive alternatives to traditional supervised algorithms, while being fully automated~\cite{vo2024automatic}. 
In contrast, data curation in \gls{dp} largely relies on expert annotations that provide high-level information at the \gls{wsi} level (\emph{e.g.,}, cancer and tissue types~\cite{hosseini2019atlas}), while potentially overlooking tile-level granularity. To this end, a semi-automatic strategy has been proposed recently, combining WSI labels and weak tile-level supervision based on color statistics~\cite{dippel2024rudolfv}. Despite this progress, the need for expert intervention still limits scalability, underscoring the necessity of a fully automated data curation framework tailored to the heterogeneity and annotation challenges of \gls{dp}.

\begin{figure}[t!]
    \centering
    \includegraphics[width=\textwidth]{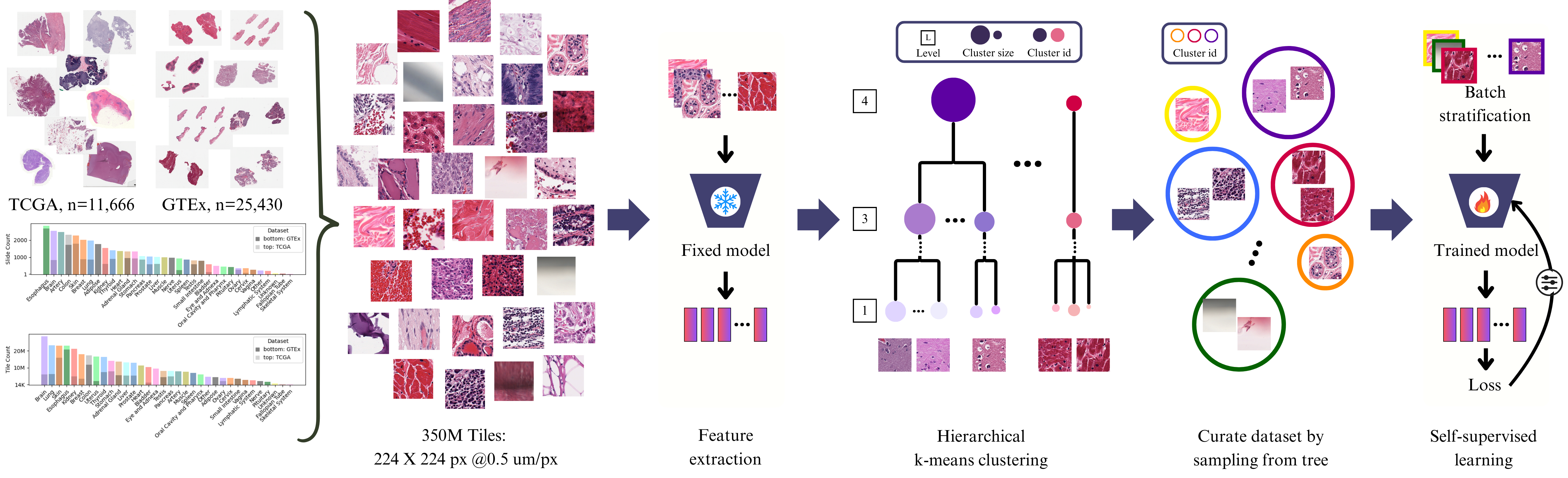}
    \caption{
    Overview of our pipeline for automatic data curation and SSL training in DP.
    \label{fig:pipeline}}
\end{figure}

In this work, we fill this gap by studying unsupervised automatic data curation for \gls{dp} using a large set of 350 million tiles. Inspired by~\cite{vo2024automatic}, we employ \emph{hierarchical clustering trees} on tile embeddings extracted using existing \glspl{fm}. These hierarchical clusters enable uniform coverage of the data distribution across the embedding space and facilitate efficient sampling of diverse, curated datasets, while allowing control over their balance and size.
This approach mitigates biases stemming from over- or under-represented patterns and permits more efficient selection of representative samples. However, our empirical results indicate that directly applying this curation method in \gls{dp} does not consistently enhance the discriminative power of learned \gls{fm} embeddings compared to using uncurated data. We identify the data feeding strategy of the \gls{ssl} model in~\cite{oquab2023dinov2} as the limiting factor and address it by introducing batch stratification based on hierarchical clusters, which forces models to continuously learn how to represent more diverse samples while mitigating the biases resulting of the imbalance present in our heavy-tailed data distribution. Our full pipeline is depicted in Figure~\ref{fig:pipeline}. 
To summarize, our contributions are three-fold:
1) we present, to the best of our knowledge, the first \emph{fully automated} data curation scheme for \gls{fm} training in \gls{dp};
2) we identify batch stratification, alongside hierarchical clustering-based curation, as the key to data distribution uniformization and improved downstream performance;
3) we demonstrate the effectiveness of our method through improved performance on both \gls{roi}- and \gls{wsi}-level benchmarks.
Our code and curated data can be found on GitHub\footnote{https://github.com/swiss-ai/patho-ssl-data-curation}.

\section{Method}
In this section, we describe our automatic data curation framework for training \glspl{fm} in \gls{dp}. We first detail the pre-training dataset and the curation procedure, and then explain how to effectively utilize the curated data for \gls{fm} training.

\subsection{Hierarchical Clustering for Automatic Data Curation}\label{sec:data_curation}

We build a large pre-training dataset by collecting 11,666 \glspl{wsi} from The Cancer Genome Atlas (TCGA) cohortspanning 32 cancer types, and 25,430 \glspl{wsi} from the Genotype-Tissue Expression (GTEx, v7) dataset across 40 healthy tissue sites. Following~\cite{Campanella2019-wg}, we preprocess \glspl{wsi} at $\times$20 magnification (\emph{i.e.}, 0.5 $\mu$m/pixel) by first detecting tissue regions and extracting $\sim$350 million non-overlapping $224\times 224$ pixel tiles. Tissue-specific distributions are shown in Figure~\ref{fig:pipeline}. 

These two steps define a recursion applied until reaching the bottom-level clusters, where the allocated points are \emph{sampled randomly}.
Inspired by~\cite{vo2024automatic}, we first extract tile embeddings from the pre-training dataset using a well-established \gls{fm} in \gls{dp} (\emph{i.e.}, UNI~\cite{chen2024uni}) and construct a hierarchical clustering tree in a bottom-up manner. At the bottom level, K-means is applied to all tile embeddings, generating more clusters and of relatively smaller volumes in dense areas than in sparse ones. Data imbalance is artificially reduced, while the average nearest-neighbor distance is increased, by selecting the cluster centroids as the new data distribution. We recursively apply K-means to these centroids with an exponentially decreasing number of clusters, ensuring more uniform cluster volumes at each hierarchical level. This recursion provably yields \emph{clusters that distribute more uniformly over the data support as one ascends the hierarchy}~\cite{oquab2023dinov2}. Remark that both K-means++ and the resampling strategies proposed in~\cite{vo2024automatic} are utilized in practice to improve the final hierarchical clustering. Next, a top-down sampling strategy is applied to sample curated subsets of size $N$. First, sample sizes are allocated to the $k$ top-level clusters $\{c_i\}_{i=1}^k$ of respective total sizes $\{s_i\}^k_{i =1}$, by finding the size $n$ which minimizes \(|N-\sum_{i}^{k}min(n, s_i)|\) through a binary search in \(\{0, \ldots,N\}\). This process is then repeated for each cluster $c_i$ independently, distributing $min(n, s_i)$ samples across its respective downstream clusters. 
Overall, this method allows us to efficiently sample subsets of great diversity, while controlling the trade-off between balance and size.

Since no principled method exists to select an optimal size for a given tree configuration, nor its depth and width that influence this trade-off, we investigate two tree configurations across various subset sizes. First, following~\cite{vo2024automatic}, we set the depth to 4 and the number of leaves to 1\% of our dataset. Then, we select 62 and 2048 as the top-level cluster counts, yielding the following cluster distributions per tree level: $\textbf{T1}:\{3.5M, 35k, 350, 62\}$ and $\textbf{T2}:\{3.5M, 100k, 10k, 2048\}$. 
\begin{figure}[t!]
    \centering
\includegraphics[width=0.8\linewidth]{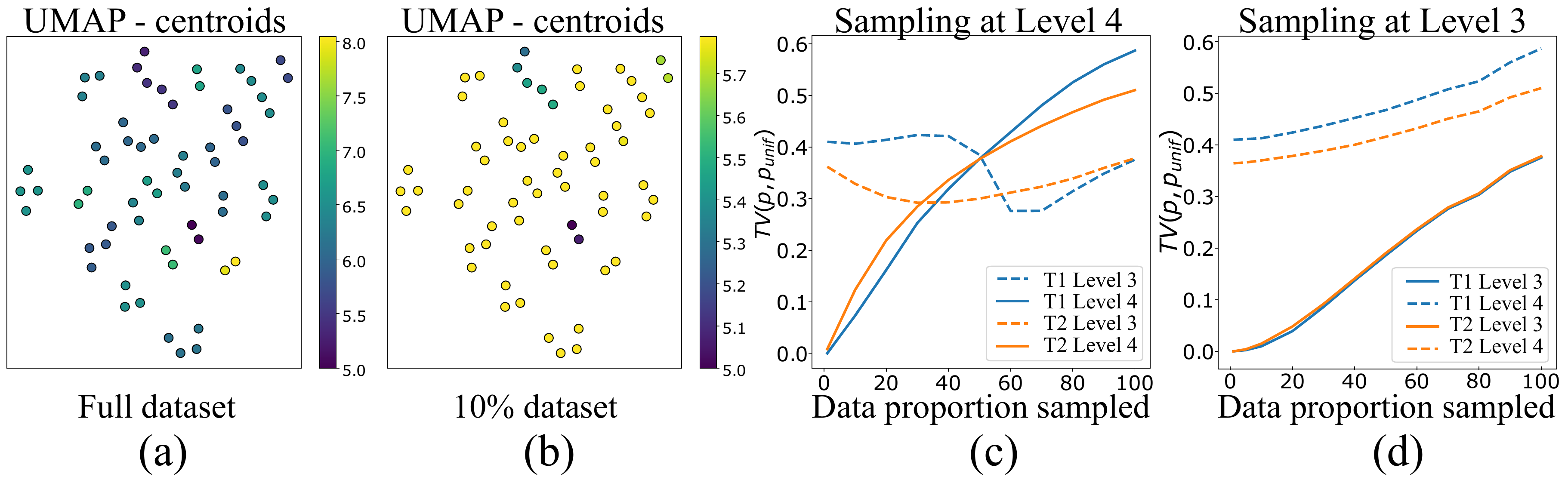}
    \caption{
    (a, b) UMAP embeddings of T1 centroids at level 4. Colors represent cluster sizes (logarithmic scale) for the full and the 10\% curated dataset, respectively. (c, d) Visualization of hierarchical sampling dynamics for T1 and T2. Solid lines denote the sampling level and dashed lines are the corresponding dynamics at the alternative level.
    } 
    \label{fig:visu_trees}
\end{figure}
We note that T2 closely mirrors the hyperparameters in~\cite{vo2024automatic}, with a higher top-level cluster count (equals to the global batch size, see Section \ref{sec:ssl_method}) and branch expansion rates, whereas T1 has fewer top-level clusters, following insights from expert annotations~\cite{hosseini2019atlas}, thereby requiring greater compression at each level. In Figure~\ref{fig:visu_trees}, we use these trees to analyze the data distribution and illustrate the effects of hierarchical sampling. Figure~\ref{fig:visu_trees}(a) presents UMAP embeddings of top-level centroids from T1, with hyperparameters chosen to best preserve local relative positions. Centroids are colored based on their effective size without sampling.
The results indicate that top-level clusters indeed cover rather uniformly our distribution and that it has heavy tails with 2 modes encompassing approximately two-thirds of the dataset. The results from sampling 10\% of the data (Figure~\ref{fig:visu_trees}(b)) show that while global diversity is preserved, the curated subset remains slightly imbalanced. We further analyze the trade-offs between balance and size via the total variation distance (TV) between sampled cluster proportions ($p$) and uniform weights ($p_{unif}$) across two clustering levels. Results shown in~Figure~\ref{fig:visu_trees}(c) and (d) reveal that the large imbalance in our dataset implies a significant imbalance for all curated subsets with more than 1\% data, since the smallest top-level clusters are depleted during sampling. We also observe that sampling from any level results in a nearly maximal imbalance at the other level studied, as densest level 4 clusters contain more clusters at level 3. This implies, for instance, that uniform sampling of a small amount of data at level 4, \emph{e.g.,} with TV near 0 at this level, propagates more samples in fewer level 3 clusters within sparser level 4 clusters, and vice versa in denser level 4 clusters, resulting in a high TV at level 3.

\subsection{Effective Self-supervised Learning on Curated Data}\label{sec:ssl_method}

After the data curation, we proceed to the self-supervised training of \glspl{fm} using the curated datasets.
We follow existing works on \glspl{fm} in \gls{dp}~\cite{vorontsov2023virchow,xu2024gigapath,chen2024uni,dippel2024rudolfv} and adopt DINOv2~\cite{oquab2023dinov2} as our \gls{ssl} framework. We utilize large Vision Transformers (ViT-L)~\cite{dosovitskiy2020image}, trained on the curated 10\% subset for 170 thousand iterations with a global batch size of 2048, equivalent to one complete pass over the full dataset (350 million tiles).
Typically, these batches are randomly sampled from the pre-training dataset (curated or not), as in most \glspl{fm} training with automatic data curation~\cite{vo2024automatic} or without it~\cite{oquab2023dinov2,vorontsov2023virchow,xu2024gigapath,chen2024uni}. Instead, we propose to structure these batches during training by stratifying them based on the tree clusters from which the tiles were sampled. 
In particular, each batch is constructed to contain an equal number of tiles from each top-level cluster, 
leveraging the inherent balancing of curation clusters to guarantee uniform coverage of the data distribution within batches. This approach ensures that differently populated clusters are equally represented during training, prompting models to continually adapt to diverse samples while mitigating biases introduced by dataset imbalance.
To further promote intra-cluster diversity in batch construction, we track the samples encountered during training and prioritize those that have been less frequently sampled: for each batch and cluster, we sample exclusively from the least observed tiles within that batch and cluster.
This procedure guarantees that every tile in the curated dataset is observed multiple times during training.

\section{Experiments}
In this section, we first examine whether the clusters identified by hierarchical clustering exhibit interpretable biological features. We then present a comprehensive benchmark of data curation and learning strategies, followed by a more detailed investigation the sensitivity of our method to its hyperpameters.\\

\noindent\textbf{Interpretability of hierarchical clusters.} To study the interpretability of hierarchical clusters, we analyze the 1\% curated subset using T1 at level 4. We first conduct visual inspections of tile images in the clusters, an example of which is shown in the supplementary video. We then follow~\cite{zhao2023single} to quantitatively report our findings, by extracting basic statistics for handcrafted single-cell features (color intensities, textures, morphologies, spatial arrangements) and including class densities from CellViT~\cite{horst2024cellvit} and extracellular matrix descriptors. We observe that $\sim$50\% highest-level clusters mainly contain tiles of healthy tissues from GTEx and a few TCGA tiles without neoplastic cells, while the two largest clusters are mainly cancerous TCGA tiles with $\sim$50\% cells neoplastic or inflammatory and the remainder epithelial or connective tissue cells.

Next, we cluster the tiles using handcrafted features by applying UMAP, followed by K-means with the same number of clusters as T1’s top level. The resulting clusters are then compared to those obtained from hierarchical clustering on raw embeddings using the Adjusted Rand Index (ARI). Our experiments show that both types of clusters match weakly with ARI scores of 7.7\% at level 4 and 5.5\% at level 3. Moreover, restricting the same process to specific types of handcrafted features reveals that cell colors and spatial arrangements are relatively the most discriminative factors across hierarchical clusters. Overall, these results suggest that while handcrafted features explain the hierarchical clusters to some extent, they do not fully capture the underlying patterns dictating this organization.

\noindent \textbf{Evaluation framework.} 
In the following, we evaluate different data curation settings at both RoI- and WSI-level. Specifically,  8 \gls{roi}-level tasks from independent cohorts are considered, including lung adenocarcinoma classification (LUAD~\cite{han2022wsss4luad}), colorectal tissue and polyp classification (CRC~\cite{kather2019predicting}, UniToPatho~\cite{barbano2021unitopatho}, Chaoyang~\cite{zhu2021hard}), breast cancer subtyping (BRACS~\cite{brancati2022bracs}) and tissue classification (BACH~\cite{aresta2019bach}, BreakHis~\cite{spanhol2015dataset}), and lymph node metastatis classification (PCAM). All \gls{roi} images are resized to 224×224 pixels before embedding extraction. We perform linear probing over 1000 nonparametric bootstrap iterations on embeddings extracted from each encoder. At the WSI-level, we assess 9 clinically relevant tasks~\cite{chen2024benchmarking} from two different clinical centers (MSHS and MSKCC), including breast cancer (BCa) detection, and biomarker prediction for Estrogen Receptor (ER), Progesterone Receptor (PR) and Human Epidermal Growth Factor Receptor (HER2) and breast Homologous Repair Deficiency status (HRD). Further tasks include prediction of lung Epidermal Growth Factor Receptor (EGFR) mutation status and immunotherapy (IO) response in lung cancer patients, and, lastly, detection of inflammatory bowel disease (IBD) versus normal mucosa samples. We represent each WSI as a set of tile embeddings composing its segmented tissue regions and train attention-based multiple instance learning~\cite{ilse2018attention} models over 20 Monte Carlo cross-validation runs~\cite{ilse2018attention}. We report average balance accuracy (bACC, in \%) and area under the receiver operating characteristic curve (AUC, in \%) for \gls{roi}- and \gls{wsi}-level tasks, respectively~\cite{chen2024uni,chen2024benchmarking}.

\noindent \textbf{Benchmarked methods.} 
We compare several data curation and batch sampling strategies. The baseline method, F-BR, learns from the \emph{full dataset} with random batch sampling.
We also include a \emph{supervised data curation} approach where \glspl{wsi} are first split into 266 classes, with healthy tissue types for GTEx and combinations of tissue type and primary cancer diagnosis for TCGA, before sampling tiles evenly across classes. Two methods, denoted S-BR and S-BS, learn from these curated datasets using random and stratified batch sampling, respectively. Finally, for the \emph{automatic data curation} introduced in Section~\ref{sec:data_curation}, we select the curated tiles by sampling from T1 and T2 at level 4. For training, we employ both random and stratified batch sampling, denoting these methods as T-BR and T-BS, respectively, with $T \in \{\text{T1}, \text{T2}\}$ indicating the tree used. Unless stated otherwise, we use curated subsets with 10\% of the total data.

\noindent \textbf{Main results.}
The performance on the \gls{roi}- and \gls{wsi}-level benchmarks are presented in~Tables \ref{tab:roi} and~\ref{tab:wsi}, respectively. We can observe that our approach, T1-BS, achieves the highest average performance across both benchmarks, followed by S-BS and F-BR. In contrast, T1-BR and S-BR yield the lowest average performance. These results emphasize the effectiveness of jointly applying automatic data curation at the tile level and stratified batch sampling strategies. Nevertheless, rankings per task still vary significantly. For further analysis, we note that our benchmarks predominantly consist of images related to breast, colon, and lung tissues. Thus, we can expect any inherent bias towards these tissue types in benchmarked methods could lead to improved performance. To better disentangle model performance from such biases, we first provide a detailed breakdown of the dataset compositions used during training. In the full dataset, breast, colon, and lung tissues account for 5.2\%, 5.1\%, and 6.7\% of tiles, with 61\% coming from TCGA, which are more likely cancer tissues. 
In the supervised curation subset, these proportions shift to 6.1\%, 3.2\%, and 9.3\%, with 80\% from TCGA, dispersed across 58 of 266 classes. The hierarchical clustering subset contains 2.7\%, 3.9\%, and 3.9\%, with 53\% from TCGA, covering all level-4 clusters. From a global perspective, both T1-BS and T1-BR utilize on average the smallest proportions of tiles from these tissues while maintaining competitive performance, underscoring the effectiveness of automatic data curation in enhancing diversity in sampling. Among the most comparable settings, \emph{e.g.,} T1-BS and S-BS, we observe strong correlations between performance and both tile proportions and their origins. Both methods use similar tile proportions from GTex for breast and lung tiles, but S-BS uses increasingly more TGCA tiles for each tissue. However, T1-BS outperforms S-BS on \gls{roi}-level tasks related to breast tissue, achieving an average balanced accuracy of 87.2\% compared to 84.7\% for S-BS, while demonstrating slightly better performance on the LUAD lung dataset. Notably, T1-BS leads to the largest improvement on the most challenging task, BRACS, with 7 cancer subtypes. For colon tissues, S-BS uses fewer tiles than T1-BS yet their average performance remains comparable at 71.1\% and 70.9\%, respectively. T1-BS actually tends to outperform S-BS in benchmarks with greater tissue diversity, achieving superior performance on CRC (9 classes) while performing similarly or slightly worse on UniToPatho (6 classes) and Chaoyang (4 classes), suggesting its potential for training a better generalist \gls{fm}. 

For WSI-level tasks, we recall that AUC considers model calibration and therefore could accentuate the effects of the biases discussed above. Nonetheless, we observe similar trends in performance reinforcing that T1-BS better transfers across tasks. For breast-related biomarker prediction (ER, HER2, PR, HRD) and cancer detection (BCa), T-BS achieves slightly higher average AUC of 86.8\% and 97.6\%, compared to 86.4\% and 97.5\% for S-BS. In lung-related benchmarks (EGFR and IO), T1-BS outperforms S-BS by a larger margin of 0.9\% in average AUC. Notably, T1-BS shows greatest improvement in the two most challenging tasks: HRD and IO prediction in breast- and lung-related benchmarks. \\

\begin{table*}[t!]
  \centering
  \fontsize{6}{8}\selectfont
  \caption{\label{tab:roi}RoI-level evaluation. Best results are in \textbf{bold} and second best are \underline{underlined}.}
  \begin{tabular}{c||c|cccc|ccc|c}
    \toprule
    \multirow{2}{*}{Setting} & \multicolumn{1}{c|}{Lung} & \multicolumn{4}{c|}{Breast} & \multicolumn{3
    }{c|}{Colon} & \multirow{2}{*}{Overall} \\
    \cmidrule(lr){2-9} 
     & LUAD & BRACS & BreakHis & BACH & PCAM & CRC & UniToPatho & Chaoyang &  \\
    \midrule
    \textsc{F-br}  & \textbf{94.2\std{0.7}}  & \underline{66.1\std{2.5}} & 96.9\std{0.5}   & 85.0\std{3.9}   & \textbf{91.1}\std{0.2} & 86.8\std{0.3} & \underline{42.4\std{1.0}} & 76.7\std{1.0} & 79.9 \\
    \textsc{S-br}  & 93.7\std{0.7}   & 61.9\std{2.3}  & 96.4\std{0.6}   & 88.5\std{3.6}   & \underline{90.9\std{0.2}} & 86.8\std{0.4} & 37.6\std{1.1}  & 74.8\std{1.1} & 78.9 \\
    \textsc{S-bs}  & \underline{94.1\std{0.7}}  & 62.4\std{2.3}  & \textbf{98.3\std{0.5}} & 87.4\std{3.8}   & 90.8\std{0.2}  & \underline{90.8\std{0.3}}  & \textbf{43.7\std{1.0}}  & \textbf{78.8\std{1.0}}  & \underline{80.8} \\
    \textsc{T1-br} & 93.8\std{0.7}   & 63.8\std{2.4}  & 96.6\std{0.5}   & \underline{88.8\std{3.4}}  & 90.6\std{0.2}  & 87.4\std{0.4}  & 41.0\std{1.1}  & 76.5\std{1.0}  & 79.8 \\
    \textsc{T1-bs} & \textbf{94.2}\std{0.7}   & \textbf{69.3}\std{2.2}  & \underline{97.5\std{0.5}} & \textbf{91.2\std{3.1}}  & 90.8\std{0.2}  & \textbf{91.9\std{0.3}}  & \textbf{43.7\std{1.1}}  & \underline{77.1\std{1.0}}  & \textbf{82.0} \\
    \bottomrule
  \end{tabular}%
  \label{tab:tilelevel_eval}
\end{table*}

\begin{table*}[t!]
  \centering
  \fontsize{6}{8}\selectfont
  {
  \caption{\label{tab:wsi}WSI-level evaluation. Best results are in \textbf{bold} and second best are \underline{underlined}.}
    \begin{tabular}{c||*{5}{c}|*{2}{c}|*{2}{c}|c}
      \toprule
      \multirow{3}{*}{Setting} & \multicolumn{7}{c|}{Biomarker} & \multicolumn{2}{c|}{Detection} & \multirow{3}{*}{Overall} \\
      \cmidrule(lr){2-10}
      & \multicolumn{5}{c|}{MSHS} & \multicolumn{2}{c|}{MSKCC} & \multicolumn{2}{c|}{MSHS} \\
      \cmidrule(lr){2-10}
       & ER & HER2 &PR & HRD & EGFR & EGFR & IO & BCa & IBD & \\
      \midrule
       \textsc{F-br}
        & \underline{96.5\std{0.8}} & 80.5\std{2.8} & \underline{91.3\std{1.2}} & 74.4\std{12.0} & \textbf{73.3}\std{5.8} & \textbf{76.2\std{3.4}} & 57.3\std{4.8} & \textbf{97.6\std{0.8}} & 96.5\std{1.0} & 82.6 \\[0.8ex]
       \textsc{S-br}
        & 96.1\std{0.8} & 79.6\std{2.7} & 91.1\std{1.5} & 72.8\std{11.3} & 71.6\std{5.2} & 75.3\std{2.9} & 56.7\std{5.9} & 97.2\std{0.8} & 96.1\std{1.2} & 81.8 \\
       \textsc{S-bs}
        & \textbf{96.6\std{0.9}} & \textbf{81.0\std{3.2}} & \textbf{91.7\std{0.9}} & \underline{76.2\std{13.7}} & 72.0\std{3.1} & 75.3\std{3.0} & \underline{57.9\std{6.3}} & \underline{97.5\std{0.8}} & \textbf{97.0\std{0.8}} & \underline{82.8} \\
       \textsc{T1-br}
        & 95.7\std{1.1} & 80.1\std{2.2} & 89.9\std{1.2} & \underline{76.2\std{10.1}} & 70.1\std{5.2} & 75.8\std{3.2} & 52.9\std{7.0} & 97.1\std{0.9} & 95.9\std{1.1} & 81.5 \\[0.8ex]
       \textsc{T1-bs}
        & 96.2\std{0.9} & \underline{80.8\std{3.1}} & 91.2\std{1.1} & \textbf{79.1\std{9.0}} & \underline{73.0\std{4.7}} & \underline{75.9\std{3.0}} & \textbf{59.0\std{5.1}} & \textbf{97.6\std{0.8}} & \underline{96.7\std{1.0}} & \textbf{83.3} \\[0.8ex]
      \bottomrule
    \end{tabular}%
  }
  \label{tab:wsilevel_eval_config}
\end{table*}

\begin{figure}[t!]    \begin{center}\includegraphics[width=0.8\textwidth]{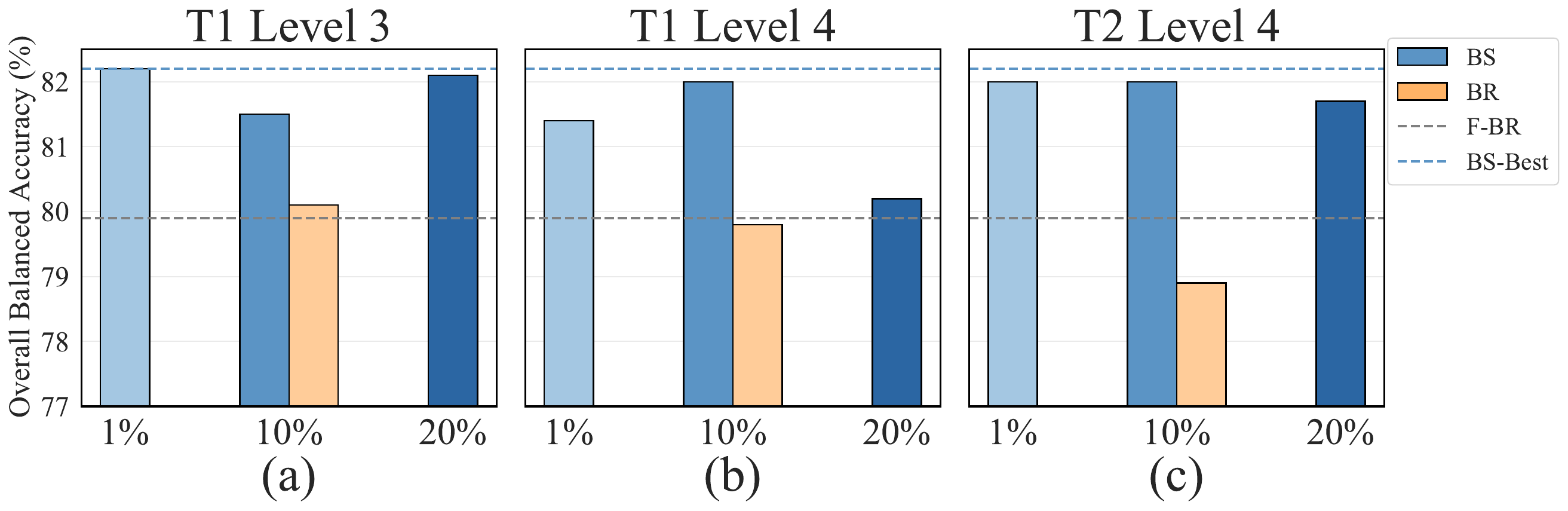}
    \end{center}
    \caption{\gls{roi}-level performances across clustering trees and batch sampling strategies.}
    \label{fig: ablation}
\end{figure}

\noindent \textbf{Sensitivity analysis.} We analyze the sensitivity of both T-BS and T-BR to sampling strategies. Figure~\ref{fig: ablation} reports the RoI-level performance of models trained on subsets curated using different trees, sampling levels and data proportions. First, with 10\% subsets from different (sub-)trees, our stratified batch sampling consistently outperforms the random one. Focusing on T-BS, top performances remain comparable across trees, with 82.2\% average bACC when sampling from T1 at level 3 and 82.0\% otherwise. However, for a given tree, performance appears sensitive to the subset sizes, particularly with fewer clusters (T1) for batch stratification. We observe that this sensitivity stems from the alignment of tissue proportions between the pre-training curated subsets and those in the downstream evaluation datasets. 
Hence, improving \glspl{fm} as generalist models motivates sampling strategies that maximize both diversity and balance, with narrower trees offering easier control as their tissue proportions converge more slowly to the full dataset. Optimizing this trade-off can be achieved with a minimal subset size by allocating samples accordingly to the volume of each bottom-level cluster (clusters covering smaller volumes get fewer samples), as attained by uniform sampling from T1 at level 4. Additionally, since the relative positions of embeddings dictate the cluster coverage of the data support, more informative bottom-level cluster sampling (beyond random) and batch construction (beyond top-level cluster-based) are promising directions for future work.

\section{Conclusion}
We investigate automatic data curation for \gls{fm} training in \gls{dp}, and propose a tailored batch sampling strategy to better convey to models the data diversity captured by hierarchical clusters and mitigate biases from data imbalance. Our results demonstrate the effectiveness of our method and highlight the relevance of data heterogeneity in \gls{dp}. 
The study of our method is based on the assumption that both data curation and batch sampling leverage UNI's informative embeddings. However, the confidence attributed to their relative positions is low, given that sampling within bottom-level clusters is random and batch stratification does not take into account the full tree hierarchy. We plan to further investigate these effects using different pre-trained \glspl{fm}, additional tree configurations and more informative sampling strategies for both curation and learning. Finally, we intend to study in more detail the transformations that our approach implies in the embedding and its applicability to continual \gls{ssl}.

\begin{credits}
\subsubsection{\ackname} 
We gratefully acknowledge funding from the ETH AI Center, the Idiap Research Institute, and the Swiss Federal Institutes of Technology strategic focus on personalized health and related technologies, as well as the EPFL and ETH core funding. We 
acknowledge funding from the Novartis Foundation for Medical-Biological Research (grant no. 22B104) and from the Swiss AI Initiative through a grant by the Swiss National Supercomputing Centre (CSCS) under project ID a02 on Alps.

\subsubsection{\discintname}
The authors have no competing interests to declare that are relevant to this article.
\end{credits}

%
%
%
\bibliographystyle{splncs04}
\bibliography{Paper-1975}
\end{document}